# Primary Care Diagnoses as a Reliable Predictor for Orthopedic Surgical Interventions

Khushboo Verma[1], Alan Michels[2,3,4], Ergi Gumusaneli[5], Shilpa Chitnis[6], Smita Sinha Kumar[7], Christopher Thompson[8], Lena Esmail[9], Guruprasath Srinivasan[9], Chandini Panchada[9], Sushovan Guha[9,10,11], Satwant Kumar[9]*

[1] Department of Neurology & Neurological Sciences, Stanford University, Palo Alto, California, USA.
[2] Department of Internal Medicine, University of Arizona, Phoenix, Arizona, USA.
[3] Rural Health Fellowship Program, Northern Arizona University, Flagstaff, Arizona, USA.
[4] Ponderosa Family Care, Payson, Arizona, USA.
[5] Department of Psychiatry, University of Arizona, Phoenix, Arizona, USA.
[6] Department of Internal Medicine, University of Texas Health Science Center, Tyler, Texas, USA
[7] Mid-Atlantic Permanente Medical Group, Washington, DC, USA
[8] Surgical Services, University of Texas Health Science Center, Tyler, Texas, USA
[9] Neuroreef Labs, Mountain View, California, USA
[10] Department of Surgery, McGovern Medical School and The University of Texas Health Science Center, Houston, Texas, USA
[11] Houston Regional Gastroenterology Institute, Houston, Texas, USA

*Corresponding author: Satwant Kumar, SatwantKumar@NeuroReef.com

**Abstract:**

*Objective*: Referral workflow inefficiencies, including misaligned referrals and delays, contribute to suboptimal patient outcomes and higher healthcare costs. In this study, we investigated the possibility of predicting procedural needs based on primary care diagnostic entries, thereby improving referral accuracy, streamlining workflows, and providing better care to patients.

*Methods*: A de-identified dataset of 2,086 orthopedic referrals from the University of Texas Health at Tyler was analyzed using machine learning models built on Base General Embeddings (BGE) for semantic extraction. Model performance was assessed using the area under the receiver operating characteristic curve (ROC-AUC), Precision-Recall Curve (PR-AUC), and Matthews Correlation Coefficient (MCC). To ensure real-world applicability, noise tolerance experiments were conducted, and oversampling techniques were employed to mitigate class imbalance.

*Results*: The selected optimum and parsimonious embedding model demonstrated high predictive accuracy (ROC-AUC: 0.874, MCC: 0.540), effectively distinguishing patients requiring surgical intervention. Dimensionality reduction techniques confirmed the model's ability to capture meaningful clinical relationships. A threshold sensitivity analysis identified an optimal decision threshold (0.30) to balance precision and recall, maximizing referral efficiency. In the predictive modeling analysis, the procedure rate increased from 11.27% to an optimal 60.1%, representing a 433% improvement with significant implications for operational efficiency and healthcare revenue.

*Conclusion*: The results of our study demonstrate that referral optimization can enhance primary and surgical care integration. Through this approach, precise and timely predictions of procedural requirements can be made, thereby minimizing delays, improving surgical planning, and reducing administrative burdens. In addition, the findings highlight the potential of clinical decision support as a scalable solution for improving patient outcomes and the efficiency of the healthcare system.

**Introduction**

The referral system can be described as the organizational structure for referring medical problems from generalists to specialists[1]. The purpose of this system is to improve patient outcomes by providing access to specialized diagnostics and treatments when the patient's medical needs exceed the expertise of general practitioners. In this system, primary care providers serve as the "gatekeepers" to specialty care providers[1]. Surgical specialties such as orthopedic surgery require seamless collaboration between primary care providers and surgical teams, for timely interventions and optimal patient outcomes. As the population ages and the need for specialized orthopedic care grows, there is an ever-growing unmet demand for orthopedic surgeons [2]. The average wait time for orthopedic surgery referrals in the US is 37.7 days [3]. However, this wait time can be significantly longer in rural settings given fewer specialists and a larger aging population[4]. It is therefore crucial for primary care providers to accurately determine which patients will benefit from orthopedic surgery referrals.

Despite the significant impact of referrals on patient care, healthcare systems, and healthcare costs, referral processes are frequently suboptimal and stochastic, emphasizing the need for augmenting and standardizing them. Referral pathways for orthopedic procedures are often negatively impacted by misaligned referrals and incomplete preoperative evaluations, likely due to the ever-growing need for orthopedic evaluations[5]. This not only delays the necessary surgical interventions resulting in poor patient outcomes, but also places considerable administrative and operational strains on healthcare systems, resulting in economic losses. Addressing these inadequacies is critical to optimize referral workflows, as well as to enhance healthcare delivery[6].

Recent advances in machine learning have emerged as transformative tools in healthcare, enabling the processing of complex clinical data, extracting actionable insights, and supporting evidence-based decision-making[7-9]. The use of natural language processing (NLP) techniques allows us to draw meaningful inferences from unstructured clinical text[10,11]. Despite these promising prospects, artificial intelligence (AI) remains underexplored in orthopedic referral systems. Therefore, in this study, we examined whether orthopedic procedural requirements can be predicted from primary care diagnostic entries. This capability can improve referral accuracy, thereby reducing administrative burdens and facilitating timely surgical interventions[5]. However, the major challenges with clinical text data include its inherent variability, noise, and context-dependence which necessitates the use of robust embedding models that can generalize[11]. Besides generalizability, the model should be capable of drawing semantic and contextual relationships from diagnostic text. In this regard, the pre-trained text embeddings, such as the Base General Embeddings (BGE) family, have demonstrated state-of-the-art performance[10,11]. Therefore, we utilized these embeddings as they are particularly well-suited to address the complexities of clinical data, thereby enabling precise identification of patients that require surgical intervention[12].

A de-identified dataset of 2,086 orthopedic referrals from the University of Texas Health at Tyler was examined to determine whether diagnostic text could predict procedural needs in orthopedic surgery. In order to enhance interpretability and performance, we assessed the machine learning models for robustness under real-world noisy conditions, data balancing, and threshold optimization. In this study, we provide the first quantitative prediction of orthopedic procedure outcomes based on referral diagnoses. Additionally, this work demonstrates that the integration of model predictions improved both procedure rates and financial outcomes. These findings underscore the potential of AI-driven referral optimization to enhance healthcare efficiency and improve patient outcomes.

**Materials and Methods:**

*Dataset*

The dataset consisted of clinical diagnostic text entries paired with procedural labels, extracted from anonymized electronic health records (EHRs) from January 4, 2024, to November 15, 2024. This study has been exempted from Institutional Review Board (IRB) review by The University of Texas at Tyler IRB (IRB 2025-031). The diagnostic entries comprised International Classification of Diseases, Tenth Revision, Clinical Modification (ICD-10-CM) codes, accompanied by their corresponding brief descriptions (Diagnoses). Each entry was assigned a binary target label: 1) *Class 0 (No Procedures)*: Diagnostic entries without associated medical procedures and 2) *Class 1 (With Procedures)*: Diagnostic entries linked to at least one medical procedure. The binary labeling system facilitated the dataset's utilization as the foundation for a classification task aimed at predicting the presence or absence of associated medical procedures based on diagnostic text. Table 1 provides a summary of the descriptive statistics for the dataset.

*Preprocessing*

To ensure the dataset's integrity and completeness for downstream analysis, missing entries were replaced with empty strings. Additionally, an enriched version of the diagnostic text, referred to as HyDE Enriched Diagnoses, was created. This enrichment was achieved using Hypothetical Document Embeddings (HyDE), wherein contextual information derived from ICD-10-CM descriptions was appended to the original diagnostic text [13]. The enriched text aimed to improve semantic representation and enhance the performance of downstream machine learning models.

Descriptive Statistics

A detailed exploration of the dataset's structure revealed key insights into its size, target label distribution, and text characteristics. Table 1 provides a summary of the descriptive statistics.

**Table 1: Dataset Descriptive Statistics**

| Statistic | Value |
| --- | --- |
| Total Datapoints | 2,086 |
| Class 0 (No Procedures) | 1,851 (88.73%) |
| Class 1 (With Procedures) | 235 (11.27%) |
| Total Characters in Diagnoses | 181,431 |
| Total Words in Diagnoses | 24,702 |
| Average Characters per Entry | 86.98 |
| Average Words per Entry | 11.84 |

The dataset exhibited a significant class imbalance, with most entries (88.73%) belonging to Class 0, compared to only 11.27% in Class 1. This imbalance posed challenges for model training and evaluation, necessitating the application of oversampling techniques, such as the Synthetic Minority Oversampling Technique (SMOTE), to address the disparity and ensure balanced representation in subsequent modeling experiments.

*Embedding Models*

To transform textual diagnostic entries into numerical representations suitable for machine learning, three pre-trained embedding models from the BGE (Base General Embeddings) family were utilized. These models, developed as part of the C-Pack project, represent state-of-the-art advancements in general-purpose text embeddings [14]. The BGE family of models has demonstrated superior performance across multiple tasks in the MTEB (Massive Text Embedding Benchmark) compared to other leading embedding models [15].

The three BGE models varied in size and dimensionality, offering different trade-offs between computational efficiency and semantic richness. Table 2 summarizes their performance on MTEB benchmark tasks, where BGE models consistently outperformed prior state-of-the-art embeddings [14].

Table 2: Performance of BGE Models on MTEB Benchmark

| Model | Dimensionality | Retrieval | STS | Pair | CLF | Re-rank | Cluster | Average |
|---|---|---|---|---|---|---|---|---|
| **BGE-small** | 384 | 51.68 | 81.59 | 84.92 | 74.14 | 58.36 | 43.82 | 62.17 |
| **BGE-base** | 768 | 53.25 | 82.5 | 86.55 | 75.53 | 58.86 | 45.77 | 63.55 |
| **BGE-large** | 1024 | 54.29 | 83.11 | 87.12 | 75.97 | 60.03 | 46.08 | 64.23 |

*Framework for Embedding Generation*

The SentenceTransformers library was used to encode the clinical text into dense vector representations [16]. For each BGE model two distinct types of embeddings were generated: 1) Base Representations: Derived directly from the raw diagnostic text (Diagnoses) and 2) Enriched Representations: Derived from the enriched diagnostic text (HyDE Enriched Diagnoses), which augmented raw text with additional contextual information. This dual encoding approach aimed to assess the impact of text enrichment on downstream classification tasks.

*Dimensionality Reduction*

To visualize the high-dimensional embeddings and assess their structural properties, Principal Component Analysis (PCA) and Uniform Manifold Approximation and Projection (UMAP) were employed. PCA was chosen for its ability to preserve global variance in the data, while UMAP was selected for its strength in capturing local neighborhood structures [17]. For UMAP, fixed implementation parameters including: a cosine metric, 15 neighbors, and a minimum distance of 0.1 were used to optimize the balance between global and local data representations. These techniques were applied to embeddings generated by BGE models, and the results were visualized as two-dimensional scatterplots, with data points colored by their binary procedural labels. PCA revealed variance-driven separability trends, while UMAP highlighted finer clustering patterns reflecting the embedding model's ability to encode meaningful semantic relationships. These methods provided interpretable insights into the embeddings' properties, with UMAP outperforming in identifying local data structures.

*Cross-Validation and Oversampling*

To ensure robust evaluation and mitigate the effects of class imbalance, a stratified 5-fold cross-validation strategy was employed, ensuring proportional representation of the minority class (Class 1: 11.27%) in both training and test splits. The Synthetic Minority Oversampling Technique (SMOTE) was applied exclusively to the training data to generate synthetic minority class samples [18]. By interpolating between existing datapoints, SMOTE enhanced the minority class representation, facilitating improved model training on underrepresented patterns without introducing noise into the evaluation process. This approach ensured a balanced and systematic assessment of model performance, critical for datasets with significant class imbalances.

*Model Training and Hyperparameter Optimization*

Random Forest classifiers were chosen as the baseline model for their robustness, interpretability, and ability to handle high-dimensional data. Hyperparameter tuning was conducted using GridSearchCV to identify optimal configurations for the number of estimators (`n_estimators` = [50, 100, 200]), maximum tree depth (`max_depth` = [None, 10, 20]), minimum samples required to split (`min_samples_split` = [2, 5, 10]), and minimum samples at leaf nodes (`min_samples_leaf` = [1, 2, 4]). The grid search utilized three-fold internal cross-validation within each training set, with ROC-AUC guiding the selection of hyperparameters. This methodology ensured that the final models were both optimized for performance and generalizable across diverse embeddings.

*Performance Evaluation Metrics*

Model performance was assessed using a comprehensive set of metrics: area under the Receiver Operating Characteristic curve (ROC-AUC), Precision-Recall Curve (PR-AUC), accuracy, and Matthews Correlation Coefficient (MCC). ROC-AUC quantified the ability to discriminate between classes, while PR-AUC captured performance under imbalanced class distributions, focusing on the precision-recall trade-off. MCC, a robust metric accounting for true and false predictions across all classes, was prioritized for its reliability in imbalanced datasets [19]. Accuracy, although commonly reported, was secondary due to its susceptibility to bias in skewed datasets. Metrics were computed for each cross-validation fold, and the mean and standard deviation were reported to capture performance variability.

*Noise Tolerance Experiment*

To investigate the robustness of embeddings to textual perturbations, a controlled noise tolerance experiment was conducted. Four noise types were applied to the diagnostic text: 1) character substitution (char_sub), 2) character deletion (char_del), 3) word swapping (word_swap), and 4) word deletion (word_del). Noise levels ranged from 0 (no noise) to 0.5 (50% perturbation), simulating scenarios of incomplete or erroneous clinical entries. For each noise type and level, embeddings were regenerated, and models were retrained and evaluated using cross-validation. This analysis provided critical insights into embedding stability under noisy conditions, highlighting the resilience of BGE embeddings, which demonstrated minimal performance degradation across word swapping and word deletion noise types.

*Balancing Techniques and Model Comparison*

Three data balancing techniques: 1) SMOTE, 2) Adaptive Synthetic Sampling Approach for Imbalanced Learning (ADASYN), and 3) random undersampling were evaluated for their efficacy in mitigating class imbalance during model training [18,20]. Concurrently, four machine learning models: 1) Random Forest, 2) Gradient Boosting, 3) Support Vector Machines (SVM), and 4) Multi-Layer Perceptron (MLP) were compared across the balanced datasets. Model performance was evaluated using mean and standard deviation of ROC-AUC, PR-AUC, accuracy, and MCC metrics across 5-fold cross-validation.

*Threshold Sensitivity Analysis*

Threshold sensitivity analysis was conducted to identify optimal decision thresholds for the Random Forest classifier. Precision, recall, and F1-scores were calculated across thresholds ranging from 0 to 1 and interpolated for consistent comparison. These metrics were averaged across cross-validation folds, enabling the generation of smooth sensitivity curves. The analysis highlighted the trade-offs between precision and recall at varying thresholds, offering actionable insights for optimizing decision-making processes in clinical contexts, where the prioritization of precision or recall depends on the specific application.

*Statistical Analysis*

Bootstrap resampling (n = 1000) was used to estimate 95% confidence intervals for performance metrics, reported as mean ± 95% CI across cross-validation folds. The significance of the difference between the two proportions was compared using the Binomial test. Pairwise differences between embedding models were assessed using the Wilcoxon signed-rank test, with p-values adjusted for multiple comparisons using the Benjamini-Hochberg method [21]. The adjusted p-values are reported as q-values. A significance threshold of $q < 0.05$ was applied. In the absence of significant differences, the most parsimonious and computationally efficient model was selected.

# Results

Based on primary care diagnostic text, this study examines advanced embedding models, dimensionality reduction techniques, and classification strategies to predict procedural needs. This work aims to improve clinical decision support, streamline clinical workflows, and provide timely and appropriate patient care by allowing accurate procedure predictions directly from diagnostic entries. In addition to highlighting the effectiveness of embedding methods and noise tolerance, the results provide actionable insights for optimizing clinical decision support systems.

*Embedding Model Performance*

The predictive performance of the embedding models including BGE-small, BGE-base, and BGE-large in both Base and HyDE variations was evaluated using cross-validation. Metrics included the area under the Receiver Operating Characteristic curve (ROC-AUC) and the Matthews Correlation Coefficient (MCC), which is particularly effective for imbalanced datasets. Mean values and 95% confidence intervals (CIs) across five cross-validation folds are reported in Table

All embedding models demonstrated robust predictive capabilities, with mean ROC-AUC values ranging between 0.865 and 0.874 across variations (see Figure 1 and Table 3). The highest mean ROC-AUC was observed for BGE-small-en-v1.5 (Base) at 0.874 (95% CI: 0.835–0.913), closely followed by BGE-large-en-v1.5 (HyDE) at 0.870 (95% CI: 0.805–0.935), and BGE-base-en-v1.5 (HyDE) at 0.868 (95% CI: 0.811–0.925). However, statistical tests revealed no significant differences in performance between Base and HyDE variations within each embedding model (q-values > 0.05; Wilcoxon signed-rank test), indicating that while HyDE offers contextual enrichment, its impact on ROC-AUC may not be substantial in this setting.

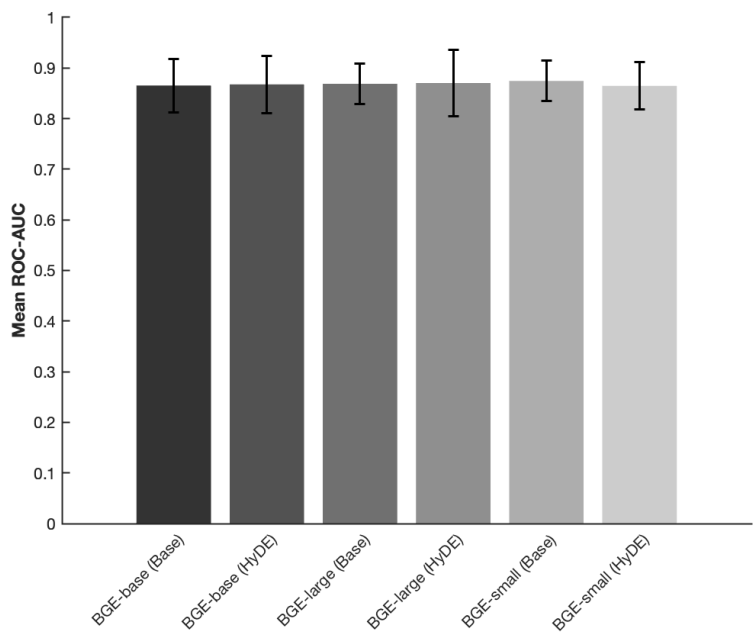

**Figure 1: Mean ROC-AUC Performance of Embedding Models**. The bar plot illustrates the mean ROC-AUC values with 95% CI error bars for the embedding models in Base and HyDE variations. Higher ROC-AUC values reflect superior predictive performance in classifying procedural requirements.

The Matthews Correlation Coefficient (MCC), summarized in Table 3, provides additional insight into the models' predictive accuracy across both classes. Among all variations, BGE-large-en-v1.5 (HyDE) achieved the highest MCC of 0.580 (95% CI: 0.492–0.668). Other models also performed well, with MCC values ranging from 0.540 (95% CI: 0.440–0.640) for BGE-small-en-v1.5 (Base) to 0.574 (95% CI: 0.478–0.670) for BGE-base-en-v1.5 (HyDE). Similar to ROC-AUC, the differences in MCC between Base and HyDE variations were not statistically significant (q-values > 0.05; Wilcoxon signed-rank test).

While HyDE variations showed slightly higher MCC and ROC-AUC values than Base counterparts, the lack of significant differences indicates comparable performance. Thus, the most parsimonious and efficient model ("BGE-small-en-v1.5 (Base)") was chosen to optimize resources without compromising accuracy. These results highlight the robustness of all evaluated models, with HyDE offering interpretative value without substantial metric improvement.

**Table 3: Evaluation Metrics for Embedding Models.** Predictive performance of embedding models across the ROC-AUC and MCC metrics.

| Model | Variation | ROC-AUC [95% CI] | MCC [95% CI] |
| --- | --- | --- | --- |
| BGE-small-en-v1.5 | Base | 0.874 [0.835 - 0.913] | 0.540 [0.440 - 0.640] |
| BGE-small-en-v1.5 | HyDE | 0.865 [0.818 - 0.912] | 0.570 [0.466 - 0.674] |
| BGE-base-en-v1.5 | Base | 0.865 [0.812 - 0.918] | 0.560 [0.476 - 0.644] |
| BGE-base-en-v1.5 | HyDE | 0.868 [0.811 - 0.925] | 0.574 [0.478 - 0.670] |
| BGE-large-en-v1.5 | Base | 0.869 [0.830 - 0.908] | 0.557 [0.453 - 0.661] |
| BGE-large-en-v1.5 | HyDE | 0.870 [0.805 - 0.935] | 0.580 [0.492 - 0.668] |

*Dimensionality Reduction*

Dimensionality reduction using Principal Component Analysis (PCA) and Uniform Manifold Approximation and Projection (UMAP) was performed to visualize the separability of binary target labels: Class 0 (No Procedures) and Class 1 (With Procedures). These techniques reduced the high-dimensional BGE-small-en-v1.5 (Base) embeddings into two dimensions for qualitative evaluation.

As shown in Figure 2, PCA revealed overlapping regions between Class 0 and Class 1, with limited clustering of Class 1 instances. This suggests that while the embeddings encode variance relevant to procedural prediction, the linear nature of PCA limits its ability to capture more complex structures. Figure 3 demonstrates that UMAP produced more distinct clusters, particularly for Class 1, with reduced but still noticeable overlap with Class 0. UMAP's ability to preserve local structures highlights non-linear relationships within the embedding space, offering better class separability than PCA.

These visualizations affirm the capacity of BGE-small-en-v1.5 embeddings to capture diagnostic features relevant to procedural prediction. The more defined clustering in UMAP suggests its utility for understanding embedding structures and supports the effectiveness of BGE embeddings for downstream classification tasks.

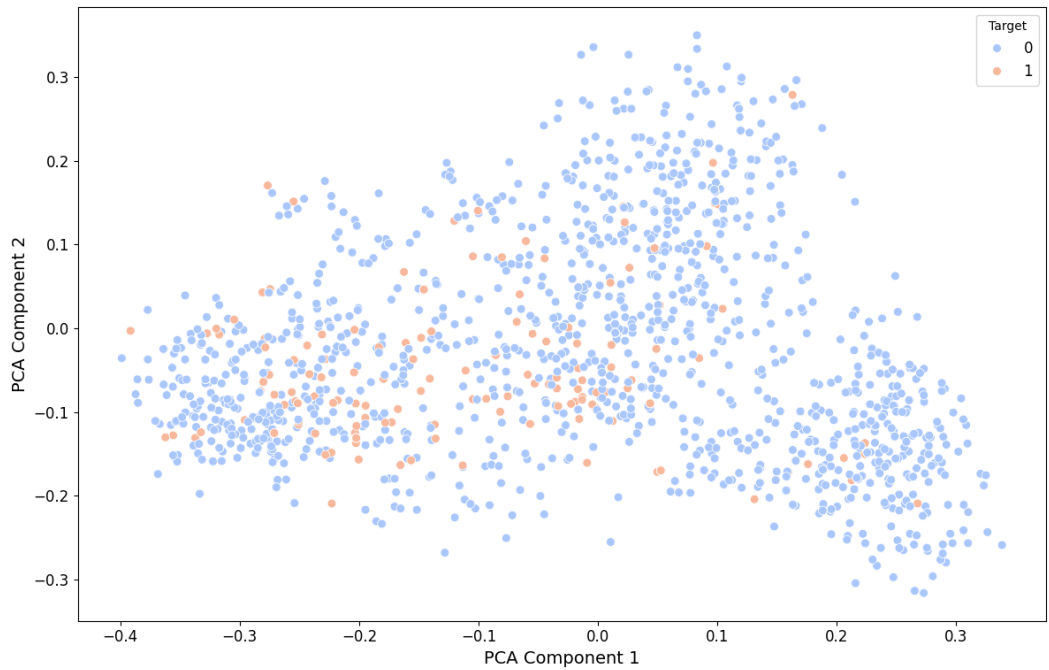

**Figure 2: PCA Visualization of Embedding Space.** Scatterplot depicting PCA-reduced embeddings with limited clustering of Class 1 (With Procedures) amidst overlapping Class 0 (No Procedures) regions.

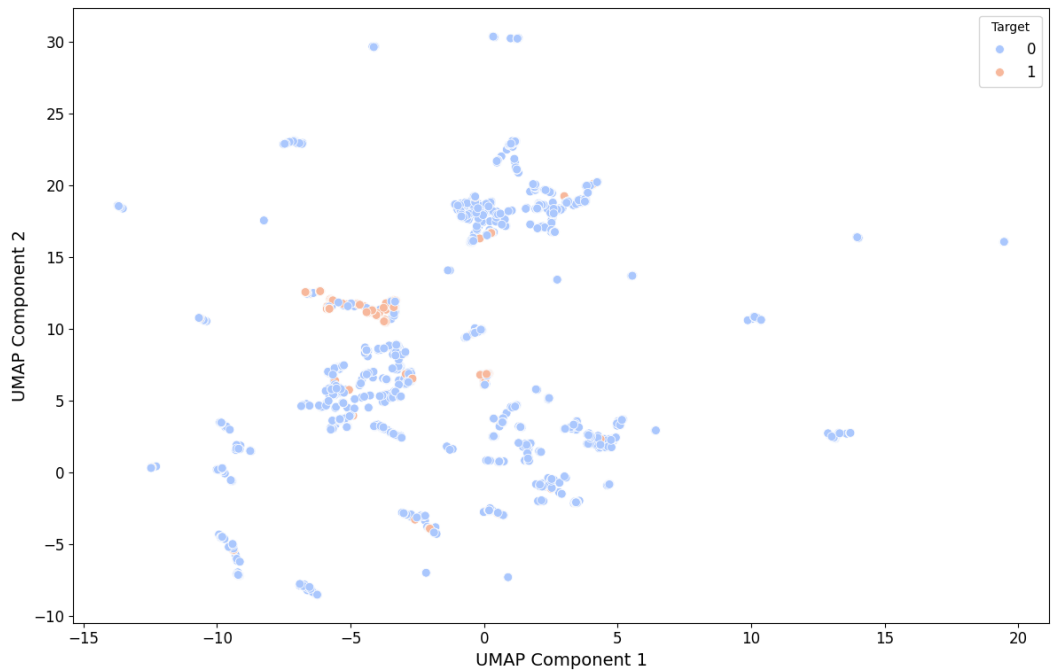

**Figure 3: UMAP Visualization of Embedding Space.** UMAP-reduced embeddings show more distinct clusters of Class 1(With Procedures), reflecting improved non-linear separability.

*Oversampling Techniques Comparison*

The impact of oversampling techniques on model performance (Random Forest) was evaluated using SMOTE, ADASYN, and undersampling. Key metrics, including ROC-AUC, Precision, Recall, F1-Score, and MCC, were calculated with 95% confidence intervals (Table 4). Both SMOTE and ADASYN significantly improved the minority class representation without compromising overall performance, achieving comparable results in terms of ROC-AUC (SMOTE: 0.863 [0.819–0.906], ADASYN: 0.863 [0.821–0.904]). Notably, ADASYN showed slightly higher recall (0.540 [0.404–0.676]) compared to SMOTE (0.532 [0.400–0.664]), while SMOTE maintained a marginally higher precision (0.667 [0.598–0.737]) and MCC (0.551 [0.448–0.654]).

In contrast, undersampling achieved the highest recall (0.800 [0.652–0.948]) at the cost of significantly reduced precision (0.355 [0.316–0.395]) and MCC (0.448 [0.355–0.540]), highlighting its trade-off between sensitivity and specificity. Overall, SMOTE was selected for subsequent analyses as it provided the best balance between predictive accuracy and minority class representation, ensuring robust performance across metrics.

**Table 4: Impact of Oversampling Techniques on Model Performance.** Performance of three data balancing techniques: SMOTE, ADASYN, and Undersampling on key evaluation metrics: ROC-AUC, Precision, Recall, F1-Score, and MCC. The results are presented as mean values with 95% confidence intervals (CI) across cross-validation folds. Techniques like SMOTE and ADASYN demonstrated balanced improvements across metrics, whereas undersampling showed a trade-off between recall and precision.

| Technique | ROC-AUC | Precision | Recall | F1-Score | MCC |
| --- | --- | --- | --- | --- | --- |
| SMOTE | 0.863 [0.819 - 0.906] | 0.667 [0.598 - 0.737] | 0.532 [0.400 - 0.664] | 0.591 [0.489 - 0.692] | 0.551 [0.448 - 0.654] |
| ADASYN | 0.863 [0.821 - 0.904] | 0.661 [0.628 - 0.693] | 0.540 [0.404 - 0.676] | 0.593 [0.503 - 0.682] | 0.552 [0.463 - 0.640] |
| Undersampling | 0.859 [0.790 - 0.929] | 0.355 [0.316 - 0.395] | 0.800 [0.652 - 0.948] | 0.492 [0.427 - 0.557] | 0.448 [0.355 - 0.540] |

*Model Comparison*

The performance of Random Forest, Gradient Boosting, Support Vector Machine (SVM), and Neural Network classifiers was evaluated using 5-fold cross-validation to identify the most effective model for predicting procedural requirements from diagnostic text. Evaluation metrics included ROC-AUC, Precision, Recall, F1-Score, and the Matthews Correlation Coefficient (MCC), with results reported as means and 95% confidence intervals (CIs) (Table 5).

Gradient Boosting achieved the highest ROC-AUC (0.870 [0.846–0.895]), followed closely by Random Forest (0.864 [0.815–0.912]). Random Forest demonstrated a balanced performance across all metrics, with the highest MCC (0.552 [0.453–0.651]) and precision (0.803 [0.688–0.917]), coupled with adequate recall (0.430 [0.339–0.520]).

Neural Network models exhibited competitive recall (0.536 [0.414–0.658]) but fell slightly behind in overall performance, as reflected in their MCC (0.549 [0.458–0.639]). In contrast, SVM displayed the lowest performance, with an ROC-AUC of 0.832 [0.743–0.922] and an MCC of 0.502 [0.390–0.613], suggesting relatively weaker discriminative ability.

Random Forest emerged as the optimal model due to its robust performance across key metrics and computational efficiency, offering a practical balance between predictive power and resource utilization. Its selection highlights the importance of considering both accuracy and scalability in deploying machine learning models for clinical decision support.

**Table 5: Performance of Machine Learning Models.** Performance metrics are presented as mean values with 95% confidence intervals, aggregated across 5-fold cross-validation for each classifier.

| Model | ROC-AUC | Precision | Recall | F1-Score | MCC |
| --- | --- | --- | --- | --- | --- |
| Random Forest | 0.864 [0.815–0.912] | 0.803 [0.688–0.917] | 0.430 [0.339–0.520] | 0.559 [0.463–0.655] | 0.552 [0.453–0.651] |
| Gradient Boosting | 0.870 [0.846–0.895] | 0.706 [0.634–0.777] | 0.404 [0.299–0.509] | 0.512 [0.422–0.601] | 0.491 [0.412–0.570] |
| SVM | 0.832 [0.743–0.922] | 0.768 [0.647–0.889] | 0.379 [0.265–0.492] | 0.505 [0.390–0.620] | 0.502 [0.390–0.613] |
| Neural Network | 0.852 [0.794–0.910] | 0.661 [0.578–0.744] | 0.536 [0.414–0.658] | 0.590 [0.503–0.677] | 0.549 [0.458–0.639] |

*Noise Tolerance Experiment*

To assess the robustness of embedding (BGE-small-en-v1.5) to noise, we introduced systematic perturbations in the input text across four noise types: character substitution (char_sub), character deletion (char_del), word swapping (word_swap), and word deletion (word_del). Noise levels ranged from 0% to 50%, and the predictive performance was quantified using the ROC-AUC metric. The results, summarized in Figure 4, demonstrate the embeddings' varying degrees of resilience to these perturbations.

Overall, embeddings were most robust at low noise levels (≤10%), maintaining mean ROC-AUC values above 0.85 across all noise types. However, the performance degraded substantially as noise levels increased, particularly for character-based perturbations. At 50% noise, character substitution exhibited the steepest decline, with a mean ROC-AUC of 0.582 [0.537–0.627], underscoring the susceptibility of embeddings to frequent character-level distortions. Similarly, character deletion led to significant performance losses, albeit slightly less severe than character substitution. In contrast, word-based noise types (word swapping and word deletion) demonstrated greater tolerance, with word swapping achieving the highest robustness, retaining a mean ROC-AUC of 0.861 [0.804–0.918] even at the highest noise level.

These findings highlight the critical importance of noise-tolerant embeddings for clinical text applications, where real-world data often contain typographical errors or omissions. While character-level perturbations severely disrupted predictive performance, word-based noise had a comparatively moderate impact, making BGE embeddings particularly well-suited for noisy environments. This resilience is vital for ensuring the reliability of predictive models in healthcare contexts, where noisy input data is inevitable.

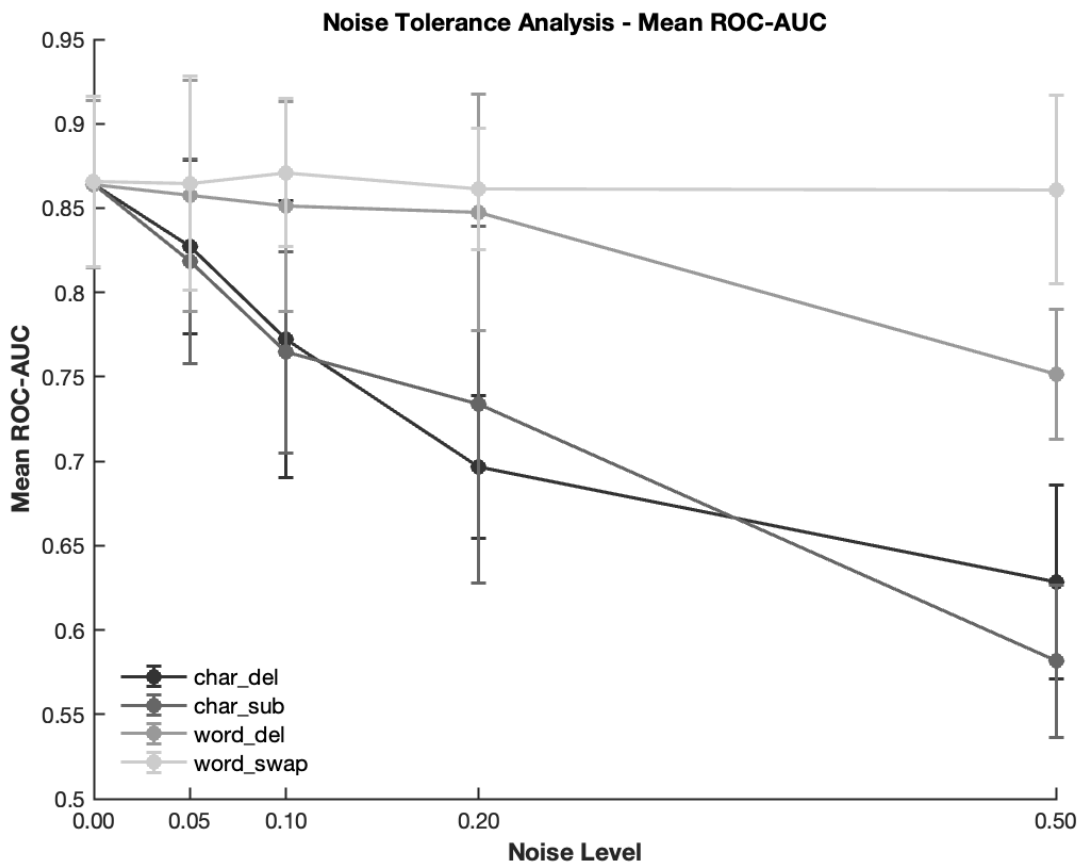

**Figure 4: Noise Tolerance Analysis - Mean ROC-AUC.** The line plot illustrates the impact of increasing noise levels on the mean ROC-AUC values of BGE embeddings subjected to four noise types. Word-based perturbations (word_swap and word_del) demonstrated greater robustness compared to character-based perturbations (char_sub and char_del). Word swapping (word_swap) maintained the most stable performance across all noise levels. Error bars represent 95% confidence intervals.

*Threshold Sensitivity Analysis*

The threshold sensitivity analysis evaluated the interplay between precision, recall, and F1-score across varying decision thresholds (0–1). As illustrated in Figure 5, these metrics demonstrated inverse relationships, with precision increasing at higher thresholds and recall exhibiting a declining trend. F1-score, a harmonic mean of precision and recall, provided a balanced measure of the model's performance and peaked at an optimal threshold of 0.30.

At lower thresholds, the model achieved high recall, ensuring that most true positives were identified. However, this came at the expense of precision due to the inclusion of false positives. In contrast, higher thresholds prioritized precision by limiting predictions to cases with higher confidence, thereby reducing recall. The F1-score reached its maximum at a threshold of 0.30, marking the point of equilibrium between precision and recall. This threshold represents the most effective compromise for maximizing overall model performance in imbalanced classification settings.

The optimal threshold of 0.30 holds significant implications for clinical decision-making. In contexts such as predicting procedural requirements, a balanced approach is essential to minimize false positives while maintaining high sensitivity to identify true cases. The identified threshold ensures that the model can provide reliable predictions without overburdening healthcare systems with unnecessary referrals or missing critical cases. This adaptability underscores the model's utility in real-world clinical applications.

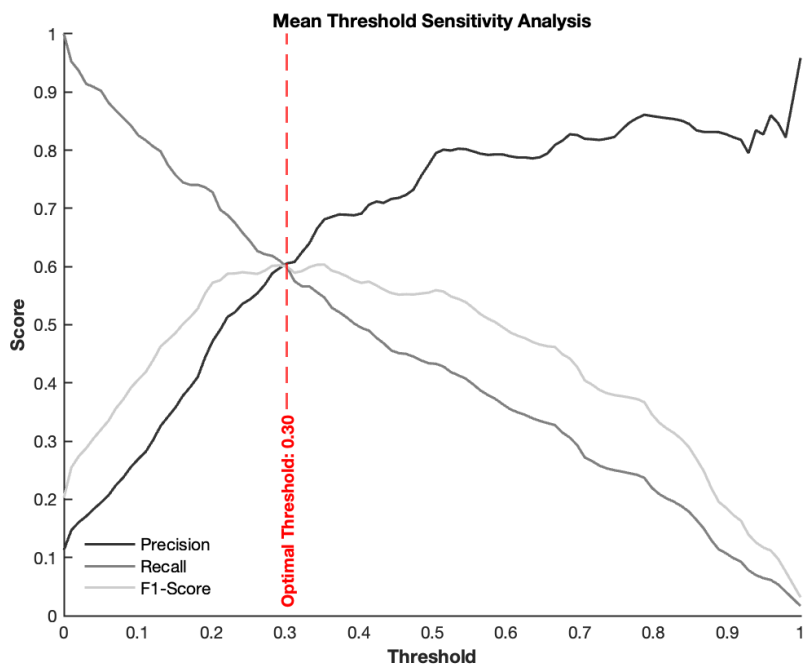

**Figure 5: Mean Threshold Sensitivity Analysis.** The line plot displays precision, recall, and F1-score across decision thresholds. The red dashed line indicates the optimal threshold of 0.30, where F1-score reached its maximum, reflecting a balanced trade-off between precision and recall. This threshold is critical for optimizing the model's performance in clinical decision-making scenarios.

*Impact on Procedure Rate Improvements and Capture Efficiency*

Our predictive modeling approach showed significant potential for improving procedure rates. The optimal procedure rate, calculated using model predictions based on primary care diagnostic entries, is 60.1%, compared with the current 11.27% procedure rate calculated from data. In terms of relative increase, this represents a substantial 433% increase. Statistically significant differences between the current and optimal procedure rates were confirmed by a two-proportion binomial test ($p < 0.001$). These findings emphasize the capacity of predictive models to improve healthcare efficiency by optimizing referral workflows in orthopedic surgery.

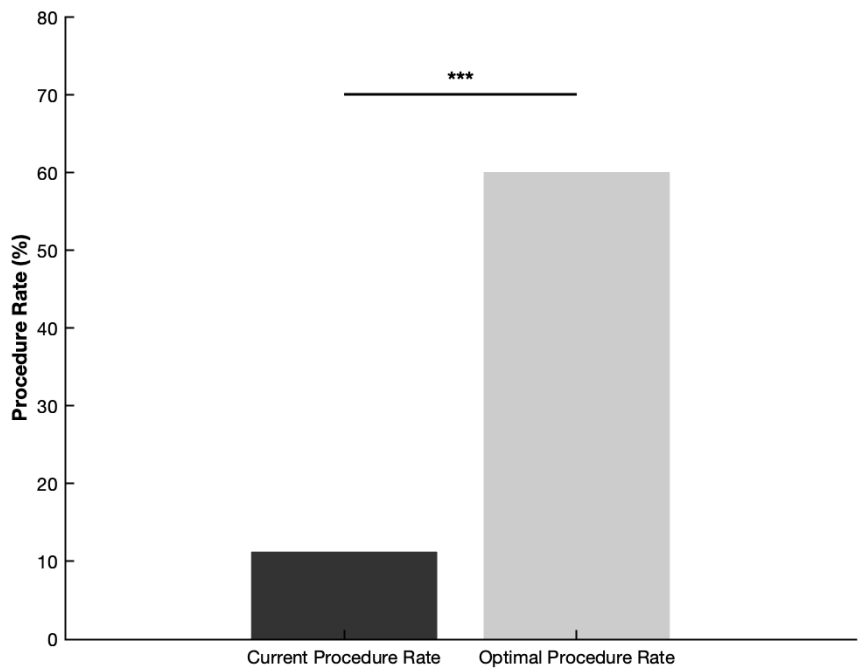

**Figure 6. Impact of Model Predictions on Procedure Rates.** Bar plot illustrating the current procedure rate (11.27%) and the optimal procedure rate (60.1%) achieved using model predictions based on primary care diagnostic entries. The optimal procedure rate reflects a 433% increase relative to the current rate. A statistical comparison of the two rates using a two-proportion binomial test confirmed a significant difference (*** = $p < 0.001$). The significance annotation is displayed above the bars.

The impact of predictive modeling was further analyzed as a function of capture efficiency, representing the proportion of model-predicted referrals incorporated into the healthcare system. Using varying capture efficiency levels, the model's potential impact on procedure rates and healthcare revenue was tested by simulating 5000 referrals to the UT Health Tyler Orthopedics services and assessing the proportion of predicted referrals successfully integrated into the healthcare system (Table 6). The baseline procedure rate of 11.27% served as the reference point, while the optimal rate predicted by the model was 60.1%. At 5% capture efficiency, the effective procedure rate increased to 13.33%, representing an 18.27% improvement from baseline. A conservative assumption of $5000 per outpatient procedure based on the CMS price estimation tool (https://www.medicare.gov/procedure-price-lookup/; accessed January 15th, 2024) resulted in 667 procedures and $1.67 million in increased revenue. Progressively higher capture efficiencies led to exponential improvements. At 40% capture efficiency, the effective procedure rate reached 27.80%, a 146.68% increase from the baseline, corresponding to 1,390 procedures performed and an additional $18.63 million in revenue. At 80% capture efficiency, the effective procedure rate peaked at 41.09%, a 264.83% increase from baseline, with 2,054 procedures performed and a $36.83 million revenue gain.

These findings highlight the transformative potential of predictive modeling to improve procedural outcomes and financial performance. Using model predictions (e.g., 10% capture efficiency) can yield a 36.64% improvement in procedure rate, resulting in significant resource utilization, operational efficiency, and economic benefits. Such results underscore the utility of leveraging AI-driven referral optimization to enhance healthcare system efficiency and patient outcomes.

**Table 6. Impact of Capture Efficiency on Effective Procedure Rates, Percentage Increase, and Revenue Gains.** For a simulated 5000 referrals, this table illustrates the effective procedure rates, the percentage increase from the baseline procedure rate (11.27%), procedures performed, and the resulting revenue increase. Capture efficiency represents the proportion of predicted referrals successfully integrated into the healthcare system.

| Capture Efficiency (%) | Effective Procedure Rate (%) | Percentage Increase from Baseline (%) | Procedures Performed | Increase in Revenue ($, in Millions) |
|---|---|---|---|---|
| 5 | 13.33 | 18.27 | 667 | 1.67 |
| 10 | 15.4 | 36.64 | 770 | 4.03 |
| 20 | 19.53 | 73.34 | 976 | 8.75 |
| 40 | 27.8 | 146.68 | 1,390 | 18.63 |
| 80 | 41.09 | 264.83 | 2,054 | 36.83 |

## Discussion

This study highlights the transformative potential of AI in optimizing orthopedic referral workflows through the prediction of procedural requirements directly from primary care diagnoses. Utilizing a real-world, de-identified dataset of 2,086 orthopedic referrals from the University of Texas Health at Tyler, this research emphasizes the applicability of pre-trained text embeddings, robust machine learning algorithms, and noise-tolerant models in predicting optimum clinical referral pathways.

This research advances the application of AI in healthcare by addressing the underexplored domain of procedural predictions in orthopedic referrals. While prior studies predominantly focus on diagnostic predictions or treatment recommendations [12,22,23], this work bridges the gap between diagnosis and intervention, providing actionable insights to optimize orthopedic surgical workflows. As discussed previously, orthopedic surgery is a highly limited resource, and this work illustrates AI's role in optimizing the utilization of this critical resource. As illustrated in Table 6, by predicting whether a patient will undergo a procedure based on a primary care referral ICD-10 codes and descriptions, the proposed model not only improves appropriate patient-physician allocation but also increases revenue for healthcare organizations.

In this study, the BGE-small-en-v1.5 (Base) model was identified as the optimal embedding model for procedural predictions based on diagnostic texts. Moreover, the unique contribution of this study is the evaluation of this embedding model's performance under simulated noise conditions. In line with previous studies that found that pre-trained embeddings such as ClinicalBERT and BioBERT are effective in capturing nuanced linguistic patterns in healthcare texts, our study demonstrates that BGE embeddings can accurately represent semantic and contextual representations in clinical diagnosis text [8,10,11,24]. Moreover, in this study, we examined the embedding model's performance under simulated noise conditions. Noise is common in medical documentation and is most commonly observed in the form of spelling or word substitution errors. These occur frequently in free text entries such as notes [25]. A study found one spelling error for every five sentences in discharge summaries and surgical reports, and the error rate rose to 10% in follow-up notes [25]. By examining the effects of character- and word-level perturbations, our study highlights the robustness of BGE embeddings, offering valuable insights into real-world applications where data variability is common [26].

A major challenge in adopting AI models in healthcare has been the interpretability of the results as AI models often operate as "black boxes" [27]. Therefore, we have utilized Random Forest models to align with the goal of explainable AI in healthcare. Random Forest models provide transparency and enable clinicians to critically appraise the decision-making processes, thereby fostering trust and supporting regulatory compliance [28,29]. Another major challenge associated with clinical datasets is class imbalance, which is also addressed in this research. In clinical datasets, class imbalance is a pervasive problem that has been addressed through oversampling techniques, with SMOTE emerging as the most effective approach The effectiveness of SMOTE in improving model performance corroborates previous findings [18,30], while the comparative analysis of balancing techniques provides practical insights for optimizing predictive models in healthcare settings. Through the threshold sensitivity analysis, we extend previous research on precision-recall trade-offs in diagnostic prediction [31,32]. By identifying an optimal threshold for balancing precision and recall, the study offers a pragmatic approach to resource allocation optimization and clinical decision-making in imbalanced settings [32]. Furthermore, we examined the financial and procedural implications of predictive modeling as a function of capture efficiency, which represents the proportion of model-predicted referrals that are successfully integrated into the healthcare system. By simulating 5,000 referrals to UT Health Tyler Orthopedics, the model demonstrated substantial improvements over baseline procedure rates (11.27%), with an optimal predicted procedure rate of 60.1%. Even at modest capture efficiencies, significant gains were observed; for instance, a 5% capture efficiency increased the effective procedure rate to 13.33% (18.27% improvement) and resulted in an additional $1.67 million in revenue.

*Clinical Implications*

Accurate prediction of procedural requirements can minimize care delays, optimize preoperative planning, and enhance referral accuracy [5]. The interpretability of Random Forest models ensures that predictive decisions remain transparent, aligned with clinical reasoning, and trusted by clinicians, promoting their adoption in practice. Additionally, ambient AI technologies can further enhance these benefits by reducing administrative burdens and providing real-time clinical decision support, which is especially relevant in resource-limited and high-need demographics [33]. These systems adapt to evolving healthcare needs through continuous learning, seamlessly integrating into workflows to improve patient care efficiency and outcomes while alleviating clinician workload [33].

Additionally, closed-loop feedback systems have demonstrated success in enhancing medical decision-making and resource efficiency. For example, telemedicine encounters with a primary care provider and co-author (AM) for back pain patients in rural Arizona initially showed only 10% of referrals leading to surgery. After targeted education of the primary care provider from the neurosurgeon on diagnostic questions and physical exam findings, referral accuracy improved dramatically, with 80–90% of cases leading to appropriate surgical interventions. This empirical observation is consistent with a systematic review of 17 studies, which indicated that structured referral sheets and the involvement of specialists in primary care providers' education were the most effective strategies for improving referrals [1]. This closed-loop approach reduces wasted referrals, expedites access to alternative treatments, and improves satisfaction among patients, families, and providers [34]. As a result of the prediction model developed in this study, primary and secondary healthcare providers can receive real-time decision support, and referral performance analytics to scale up feedback mechanisms. Additionally, this approach may apply to other subspecialties, highlighting the possibility of a broader impact [1].

*Limitations and Future Directions*

The dataset, sourced from a single healthcare system, may limit generalizability to other clinical settings. Validation using multicenter datasets with diverse demographics is essential [35]. While focused on diagnostic text, incorporating additional data, such as imaging reports, laboratory results, and patient histories, could enhance accuracy and applicability. Finally, while BGE embeddings performed robustly, fine-tuning on orthopedic-specific datasets could further improve their ability to capture domain-specific nuances, enhancing predictive performance in complex clinical scenarios [36]. Future research should explore broader applications of AI in surgical care, including predicting surgical outcomes, optimizing resource allocation, and identifying patients at risk for postoperative complications. The ethical dimensions of AI, including bias mitigation, accountability, and privacy, require ongoing attention to ensure equitable and responsible implementation [37]. Additionally, the integration of real-time feedback loops into clinical workflows can enable continuous improvement, ensuring sustained performance and adaptability to emerging clinical challenges.

*Conclusion*

By predicting surgical requirements directly from primary care diagnostic entries, this study highlights the transformative potential of AI in reshaping orthopedic referral workflows. These findings are based on robust text embeddings, noise-tolerant models, and interpretable machine learning algorithms that provide actionable insights for improving clinical decision-making, optimizing financial outcomes, and improving operational efficiency. By bridging diagnostic and procedural gaps, this study demonstrates the significance of AI as a catalyst for precision-driven, patient-centered care that enhances the efficiency and effectiveness of healthcare delivery.

**Data availability**

The data used and analyzed in the current study will be made available from the corresponding author on reasonable request.

**Acknowledgement**

We are grateful to G. Panchada, S. Machaa, S. Naragonda, and K. Reddysai for technical support.

**Ethics Declaration**

*Competing interests*
This study was funded by Neuroreef Labs Inc (Neuroreef). S.K., L.E., C.P., S.G., and G.S. are employees of Neuroreef and may own stock as part of the standard compensation package.

# REFERENCES

1. Akbari A, Mayhew A, Al-Alawi MA, et al. Interventions to improve outpatient referrals from primary care to secondary care. *Cochrane Database Syst Rev*. 2008;2008(4). doi:10.1002/14651858.CD005471.PUB2
2. Oslock WM, Satiani B, Way DP, et al. A contemporary reassessment of the US surgical workforce through 2050 predicts continued shortages and increased productivity demands. *Am J Surg*. 2022;223(1):28-35. doi:10.1016/J.AMJSURG.2021.07.033
3. Neimanis I, Gaebel K, Dickson R, et al. Referral processes and wait times in primary care. *Canadian Family Physician*. 2017;63(8).
4. Fu MC, Buerba RA, Gruskay J, Grauer JN. Longitudinal urban-rural discrepancies in the US orthopaedic surgeon workforce. *Clin Orthop Relat Res*. 2013;471(10):3074-3081. doi:10.1007/S11999-013-3131-3
5. Stanley AL, Edwards TC, Jaere MD, Lex JR, Jones GG. An automated, web-based triage tool may optimise referral pathways in elective orthopaedic surgery: A proof-of-concept study. *Digit Health*. 2023;9:20552076231152176. doi:10.1177/20552076231152177
6. Mart JPS, Goh EL, Liew I, Shah Z, Sinha J. Artificial intelligence in orthopaedics surgery: transforming technological innovation in patient care and surgical training. *Postgrad Med J*. 2023;99(1173):687-694. doi:10.1136/POSTGRADMEDJ-2022-141596
7. Meskó B, Topol EJ. The imperative for regulatory oversight of large language models (or generative AI) in healthcare. *NPJ Digit Med*. 2023;6(1). doi:10.1038/S41746-023-00873-0
8. Verma K, Moore M, Wottrich S, et al. Emulating Human Cognitive Processes for Expert-Level Medical Question-Answering with Large Language Models. Published online October 17, 2023. Accessed October 17, 2023. https://arxiv.org/abs/2310.11266v1
9. Verma K, Kumar S, Paydarfar D. Automatic Segmentation and Quantitative Assessment of Stroke Lesions on MR Images. *Diagnostics 2022, Vol 12, Page 2055*. 2022;12(9):2055. doi:10.3390/DIAGNOSTICS12092055
10. Alsentzer E, Murphy JR, Boag W, et al. Publicly Available Clinical BERT Embeddings. Published online April 6, 2019. Accessed January 6, 2025. https://arxiv.org/abs/1904.03323v3

11. Li Y, Wehbe RM, Ahmad FS, Wang H, Luo Y. A Comparative Study of Pretrained Language Models for Long Clinical Text. *J Am Med Inform Assoc*. 2023;30(2):340-347. doi:10.1093/jamia/ocac225

12. Shickel B, Tighe PJ, Bihorac A, Rashidi P. Deep EHR: A Survey of Recent Advances in Deep Learning Techniques for Electronic Health Record (EHR) Analysis. *IEEE J Biomed Health Inform*. 2018;22(5):1589-1604. doi:10.1109/JBHI.2017.2767063

13. Gao L, Ma X, Lin J, Callan J. Precise Zero-Shot Dense Retrieval without Relevance Labels. Published online December 20, 2022:1762-1777. doi:10.18653/v1/2023.acl-long.99

14. Xiao S, Liu Z, Zhang P, Muennighoff N, Lian D, Nie JY. C-Pack: Packed Resources For General Chinese Embeddings. *SIGIR 2024 - Proceedings of the 47th International ACM SIGIR Conference on Research and Development in Information Retrieval*. 2024;1:641-649. doi:10.1145/3626772.3657878

15. Muennighoff N, Tazi N, Magne L, Reimers N. MTEB: Massive Text Embedding Benchmark. *EACL 2023 - 17th Conference of the European Chapter of the Association for Computational Linguistics, Proceedings of the Conference*. Published online October 13, 2022:2006-2029. doi:10.18653/v1/2023.eacl-main.148

16. Reimers N, Gurevych I. Sentence-BERT: Sentence Embeddings using Siamese BERT-Networks. *EMNLP-IJCNLP 2019 - 2019 Conference on Empirical Methods in Natural Language Processing and 9th International Joint Conference on Natural Language Processing, Proceedings of the Conference*. Published online August 27, 2019:3982-3992. doi:10.18653/v1/d19-1410

17. McInnes L, Healy J, Saul N, Großberger L. UMAP: Uniform Manifold Approximation and Projection. *J Open Source Softw*. 2018;3(29):861. doi:10.21105/JOSS.00861

18. Chawla N V., Bowyer KW, Hall LO, Kegelmeyer WP. SMOTE: Synthetic Minority Over-sampling Technique. *Journal Of Artificial Intelligence Research*. 2011;16:321-357. doi:10.1613/jair.953

19. Chicco D, Jurman G. The Matthews correlation coefficient (MCC) should replace the ROC AUC as the standard metric for assessing binary classification. *BioData Min*. 2023;16(1):1-23. doi:10.1186/S13040-023-00322-4/FIGURES/11

20. He H, Bai Y, Garcia EA, Li S. ADASYN: Adaptive synthetic sampling approach for imbalanced learning. *Proceedings of the International Joint Conference on Neural Networks*. Published online 2008:1322-1328. doi:10.1109/IJCNN.2008.4633969
21. Benjamini Y, Hochberg Y. Controlling the False Discovery Rate: A Practical and Powerful Approach to Multiple Testing. *Journal of the Royal Statistical Society: Series B (Methodological)*. 1995;57(1):289-300. doi:10.1111/J.2517-6161.1995.TB02031.X
22. Abdel-Hafez A, Jones M, Ebrahimabadi M, et al. Artificial intelligence in medical referrals triage based on Clinical Prioritization Criteria. *Front Digit Health*. 2023;5:1192975. doi:10.3389/FDGTH.2023.1192975/BIBTEX
23. Tyler S, Olis M, Aust N, et al. Use of Artificial Intelligence in Triage in Hospital Emergency Departments: A Scoping Review. *Cureus*. 2024;16(5):e59906. doi:10.7759/CUREUS.59906
24. Lee J, Yoon W, Kim S, et al. BioBERT: a pre-trained biomedical language representation model for biomedical text mining. *Bioinformatics*. 2020;36(4):1234-1240. doi:10.1093/BIOINFORMATICS/BTZ682
25. Ruch P, Baud R, Geissbühler A. Using lexical disambiguation and named-entity recognition to improve spelling correction in the electronic patient record. *Artif Intell Med*. 2003;29(1-2):169-184. doi:10.1016/S0933-3657(03)00052-6
26. Liu F, Demosthenes P. Real-world data: a brief review of the methods, applications, challenges and opportunities. *BMC Medical Research Methodology 2022 22:1*. 2022;22(1):1-10. doi:10.1186/S12874-022-01768-6
27. Quinn TP, Jacobs S, Senadeera M, Le V, Coghlan S. The three ghosts of medical AI: Can the black-box present deliver? *Artif Intell Med*. 2022;124:102158. doi:10.1016/J.ARTMED.2021.102158
28. Rajkomar A, Oren E, Chen K, et al. Scalable and accurate deep learning with electronic health records. *npj Digital Medicine 2018 1:1*. 2018;1(1):1-10. doi:10.1038/s41746-018-0029-1
29. Tjoa E, Guan C. A Survey on Explainable Artificial Intelligence (XAI): Toward Medical XAI. *IEEE Trans Neural Netw Learn Syst*. 2021;32(11):4793-4813. doi:10.1109/TNNLS.2020.3027314

30. Haixiang G, Yijing L, Shang J, Mingyun G, Yuanyue H, Bing G. Learning from class-imbalanced data: Review of methods and applications. *Expert Syst Appl*. 2017;73:220-239. doi:10.1016/J.ESWA.2016.12.035
31. Saito T, Rehmsmeier M. The Precision-Recall Plot Is More Informative than the ROC Plot When Evaluating Binary Classifiers on Imbalanced Datasets. *PLoS One*. 2015;10(3):e0118432. doi:10.1371/JOURNAL.PONE.0118432
32. Vickers AJ, Holland F. Decision curve analysis to evaluate the clinical benefit of prediction models. *The Spine Journal*. 2021;21(10):1643-1648. doi:10.1016/J.SPINEE.2021.02.024
33. Nahar JK, Kachnowski S. Current and Potential Applications of Ambient Artificial Intelligence. *Mayo Clinic Proceedings: Digital Health*. 2023;1(3):241-246. doi:10.1016/j.mcpdig.2023.05.003
34. Chauhan BF, Jeyaraman M, Mann AS, et al. Behavior change interventions and policies influencing primary healthcare professionals' practice—an overview of reviews. *Implementation Science 2017 12:1*. 2017;12(1):1-16. doi:10.1186/S13012-016-0538-8
35. Blonde L, Khunti K, Harris SB, Meizinger C, Skolnik NS. Interpretation and Impact of Real-World Clinical Data for the Practicing Clinician. *Adv Ther*. 2018;35(11):1763-1774. doi:10.1007/S12325-018-0805-Y/TABLES/2
36. Rasmy L, Xiang Y, Xie Z, Tao C, Zhi D. Med-BERT: pretrained contextualized embeddings on large-scale structured electronic health records for disease prediction. *npj Digital Medicine 2021 4:1*. 2021;4(1):1-13. doi:10.1038/s41746-021-00455-y
37. Ueda D, Kakinuma T, Fujita S, et al. Fairness of artificial intelligence in healthcare: review and recommendations. *Japanese Journal of Radiology 2023 42:1*. 2023;42(1):3-15. doi:10.1007/S11604-023-01474-3